\documentclass{Interspeech2024}
\usepackage{subfig}
\usepackage{hyperref}

\interspeechcameraready

\title{Seamless Language Expansion: Enhancing Multilingual Mastery in Self-Supervised Models}
\name[affiliation={1}]{Jing}{Xu}
\name[affiliation={1}]{Minglin}{Wu}
\name[affiliation={1,2}]{Xixin}{Wu}
\name[affiliation={1,2}]{Helen}{Meng}

\address{
  $^1$Department of Systems Engineering and Engineering Management,\\ 
  The Chinese University of Hong Kong, Hong Kong SAR, China\\
  $^2$Centre for Perceptual and Interactive Intelligence (CPII) Limited, HKSAR, China}
\email{\{1155170427, minglinwu\}@link.cuhk.edu.hk, \{wuxx, hmmeng\}@se.cuhk.edu.hk}

\keywords{adaptation, parameter efficient, multi-lingual, language preservation, speech re-synthesis}

\begin{document}

\maketitle
\begin{abstract}
    
      Self-supervised (SSL) models have shown great performance in various downstream tasks. However, they are typically developed for limited languages, and may encounter new languages in real-world. Developing a SSL model for each new language is costly. Thus, it is vital to figure out how to efficiently adapt existed SSL models to a new language without impairing its original abilities. We propose adaptation methods which integrate LoRA to existed SSL models to extend new language. We also develop preservation strategies which include data combination and re-clustering to retain abilities on existed languages. Applied to mHuBERT, we investigate their effectiveness on speech re-synthesis task. Experiments show that our adaptation methods enable mHuBERT to be applied to a new language (Mandarin) with MOS value increased about 1.6 and the relative value of WER reduced up to 61.72\%. Also, our preservation strategies ensure that the performance on both existed and new languages remains intact\footnote{Demo page: \href{https://jingxu96.github.io/multilingual-ssl-demo/index.html}{https://jingxu96.github.io/multilingual-ssl-demo/index.html}}.

\end{abstract}

\section{Introduction}

Self-supervised pre-trained models have significantly advanced the field of speech technology, offering robust solutions for learning general representations from vast quantities of unlabeled data\cite{sslreview,wavlm,wav2vec,hubert}. By harnessing the abundant unlabeled data, these models diminish the dependency on expensive and labor-intensive labeled data. Notably, models like Wav2vec 2.0\cite{wav2vec} and HuBERT\cite{hubert} can be easily fine-tuned for a range of downstream tasks, such as automatic speech recognition (ASR), speech synthesis and speech tokenization. 

SSL models are typically developed for a limited set of languages\cite{xlsr}, but they may encounter new languages when applied in real-world scenarios. Besides, developing a new SSL model for each new language is at high cost due to its large parameters. Hence, it is vital to adapt existed SSL models to a new language. Adapting SSL models to a new language is very challenging, as phonetic and morphological structures differ largely across languages. The situation is more severe when incorporated languages and the new language are from disparate linguistic families and have completely different attributes. For example, languages encompassed in pre-training are non-tonal such as English and the target language is a tonal language like Mandarin.  The difference between tonal language and non-tonal language is huge because tonal languages distinguish lexical or grammatical meanings through pitch variations, while non-tonal languages do not. To deal with language mismatch problem, strategies such as self-supervised adaptive pre-training and language adaptors have been proposed\cite{ssladaptive,adaptiveadapter}. In \cite{ssladaptive}, unlabeled data collected from the target language of downstream tasks are leveraged to adapt the pre-trained model prior to the last fine-tuning stage. Meta-Adaptable-Adapter is proposed in \cite{adaptiveadapter} to learn task-specific adapters for feature extraction and task-independent adapters for feature combination. However, the relationship between parameter efficiency and adaptation performance remains unclear.

When adapting an existed model to new data or new tasks, there is a phenomenon called catastrophic forgetting\cite{overcomingcf} which indicates that the model tends to forget previously acquired knowledge. The underlying cause is attributed to the model's weights update during adaptation, which inadvertently overwrite antecedent learned representations. This phenomenon can markedly impede the model's performance and capacity for generalization. To mitigate catastrophic forgetting, researchers in speech area have investigated various techniques\cite{MTL,weight} to strike a balance between adapting to new tasks or data and preserving acquired knowledge.  A multi-task learning framework is advocated in \cite{MTL} where reserving original knowledge and learning new knowledge are treated as two independent tasks. Alternative methodologies involve the orthogonal modification of weights\cite{weight,weightconsolidation,rotate}. For instance, \cite{weight} introduced a continual learning algorithm known as Regularized Adaptive Weight Modification (RAWM) for fake audio detection. Nevertheless, the strategies for sustaining the knowledge of existed languages while adapting to new ones remain insufficiently explored.
\begin{figure*}[ht]
    \subfloat[]{
        \begin{minipage}[t]{0.32\linewidth}
            \centering
            \includegraphics[width=2in]{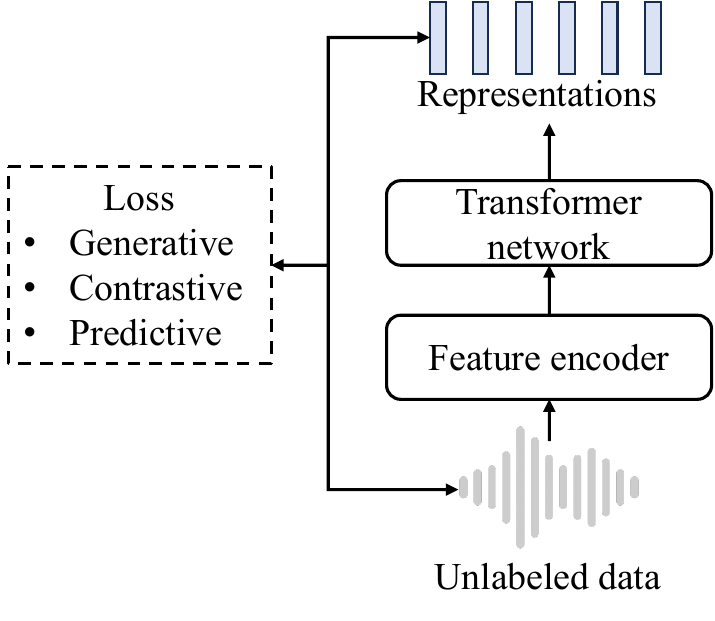}
            \label{fig1:ssl}
        \end{minipage}
    }
    \subfloat[]{
        \begin{minipage}[t]{0.32\linewidth}
            \centering
            \includegraphics[height=5cm]{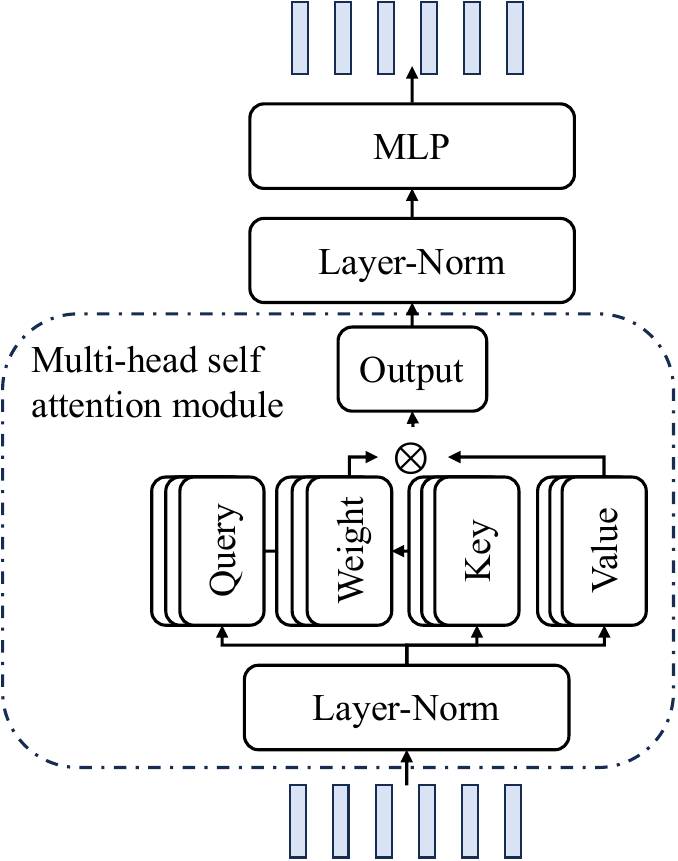}
            \label{fig1:mha}
        \end{minipage}   
    }
    \subfloat[]{
        \begin{minipage}[t]{0.32\linewidth}
            \centering
            \includegraphics[width=2in]{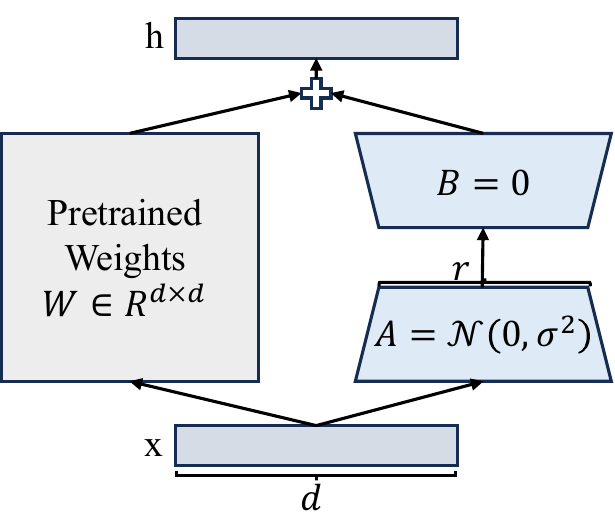}
            \label{fig1:lora}
        \end{minipage}
    }
    \vspace{-0.5em}
  \caption{(a) Architecture of self-supervised pre-trained models. (b) Architecture of Transformer blocks in self-supervised pre-trained models. (c) Architecture of LoRA which is intergrated to the self-attention module of Transformer blocks.}
  \vspace{-1.5em}
  \label{fig:PEA}
\end{figure*}

In this paper, to deal with language mismatch problem, we propose parameter efficient adaptation methods which incorporate LoRA into existed SSL models instead of developing a new one. Additionally, we develop preservation strategies that include data combination and re-clustering to mitigate catastrophic forgetting problem. They enable SSL models to be applied to new languages parameter efficiently while retaining capabilities on existed languages. Specifically, the SSL model under study is mHuBERT\cite{mhubert}, while the existed language and new language to adapt are English and Mandarin respectively. Speech re-synthesis task is adopted to evaluate the representational ability of adapted SSL model and the efficiency of adaptation. The discrete units of SSL models can naturally serve as the input of speech re-synthesis and open the potential for speech language model (SLM). Speech re-synthesis is a more difficult task than ASR since it requires both representation comprehension and waveform reconstruction. Experiments reveal that the MOS value can improve from 2.26 to 3.80 using our adaptation methods. Also, a 61.72\% relative word error rate (WER) reduction compared to un-adapted model in new language speech re-synthesis is observed. These results verify the effectiveness of our adaptation methods on adapting SSL model to a new language. Moreover, results indicate that our preservation strategies substantially mitigate the performance degradation in original languages while maintaining the newly acquired language capabilities, thereby indicating a significant mitigation of catastrophic forgetting.

\section{Method}

In this section, we first present the framework of our proposed adaptation method in subsection \ref{PEA}. Next, in subsection \ref{adapting strategies}, we illustrate two adaptation strategies while adapting self-supervised model. In subsection \ref{preserving strategies}, we propose two preservation strategies to mitigate catastrophic forgetting caused by adaptation. The evaluation system - speech re-synthesis system is introduced in subsection \ref{speech re-syn}. 

\subsection{Parameter efficient Adaptation}\label{PEA}

As shown in Figure~\ref{fig:PEA}\subref{fig1:ssl}, Transformer-based SSL models are generally composed of two principal components: a feature encoder and a Transformer network. The feature encoder, typically a deep neural network, is responsible for transforming raw audio signals into latent representations. Subsequently, these representations are input to the Transformer network, which further process them to contextualized representations. The learning paradigm of these models is multifaceted, encompassing various tasks such as generative task\cite{genemockingjay}, contrastive task\cite{wav2vec} and predictive task\cite{wavlm,hubert,data2vec}. 

Transformer network constitutes the cornerstone of SSL models' success, facilitating the capturing of intricate relationships within speech sequences. Transformer network is stacked by multiple Transformer blocks, whose architecture is shown in Figure~\ref{fig:PEA}\subref{fig1:mha}. Each block is equipped with a multi-head self-attention module, a multi-layer perceptron (MLP) layer, and layer normalization layers. Multi-head self-attention module, pivotal to the Transformer block's functionality, enables each temporal frame to compute an attention score with respect to all other frames, thereby capturing their relevance. In pursuit of efficiently enhancing multilingual mastery of SSL models, we advocate the integration of Low Rank Adaptation (LoRA)\cite{lora} into the multi-head attention module, thereby augmenting the adaptation efficiency. 

As illustrated in Figure~\ref{fig:PEA}\subref{fig1:lora}, the update of each pre-trained weight matrix $W_0\in\mathbb{R}^{d \times k}$, is constrained by representing it via a low-rank decomposition:
\begin{align}
    W_0+\Delta{W}=W_0+BA
\end{align}
where $B\in\mathbb{R}^{d \times r}$ and $A\in\mathbb{R}^{r \times k}$, and the rank $r\ll min(d,k)$.

During adaptation, only parameters of $A$ and $B$ are updated, $W_0$ are frozen, exempt from gradient updates. The inputs which multiplied with $W_0$ and $\Delta{W}=BA$ are identical, culminating the summation of their resultant output vectors on a coordinate-wise basis. Given $h_0=W_0x$, the modified forward pass is:
\begin{align}
    h = W_0x + \Delta{W}x = W_0x + BAx
\end{align}
Same as \cite{lora}, we adopt a random Gaussian initialization for $A$ and zero for $B$, so at the beginning of training $\Delta{W}=BA$ is zero. Matrix $\Delta{W}x$ is scaled by $\alpha/r$, wherein $\alpha$ is a constant in $r$. If the initialization is set appropriately, $\alpha$ is a hyper-parameter similar to the learning rate during training. Owing to the exclusive update of $A$ and $B$, the adaptation achieves great parameter efficiency. Furthermore, the stasis of matrix $W_0$ implies the preservation of the majority of the SSL parameters, thereby conferring the advanchentage of sustained performance across existed languages. 

\subsection{Adaptation strategies}\label{adapting strategies}

We propose two adaptation strategies to accomplish adaptation based on the ways of obtaining target labels. In each strategy, model parameters are initialized from an existed SSL model. In our case, the model parameters are all initialized from mHuBERT\cite{mhubert}. We will illustrate the training strategies in mHuBERT setting.
\begin{itemize}
    \item One-iteration adaptation strategy: In this strategy, the existed SSL model is directly adopted to generate continuous representations of new language. A new K-means model is trained to produce coarse target label for the new language, rendering it a time-efficient training strategy. 
    \item Two-iteration adaptation strategy: In this adaptation strategy, we iterative refine the target label. The coarse target label is derived by applying K-means on MFCC. It is used for model training in the first iteration. By applying K-means on the intermediate layer of the trained model, we can obtain fine-grained target label, which is leveraged for model training in the second iteration. To ensure a fair comparison with one-iteration adaptation strategy, the model parameters for the second iteration are initialized from publicly released model, rather than model resultant from the first iteration.
    
\end{itemize}
\subsection{Preservation strategies}\label{preserving strategies}

After adaptation, we observe the catastrophic forgetting phenomenon wherein the performance drops largely on the previously existing language. To alleviate this problem, we propose two strategies aimed at preserving the ability on existed languages of adapted model. 
\begin{itemize}
    \item Re-clustering using combined data. In this strategy, we introduce both English speech and Mandarin speech into adapted model to produce their respective representations from a designated intermediate layer. These speech samples are randomly selected from a substantial dataset, ensuring parity in the volume of English and Mandarin data. Subsequently, a K-means model is trained utilizing the combined English and Mandarin representations.
    \item Data combination in adaptation training: In this strategy, we combine English speech with Mandarin speech during adaptation training. To safeguard against a substantial diminution in performance of Mandarin, the ratio of English to Mandarin is maintained below 1. Besides, given the prior integration of English data, we limit its inclusion to a modest quantum. The K-means model is also trained on composite English and Mandarin representations.

\end{itemize}

\subsection{Speech re-synthesis system}\label{speech re-syn}

\begin{figure}[h]

  \centering
  \includegraphics[width=\linewidth]{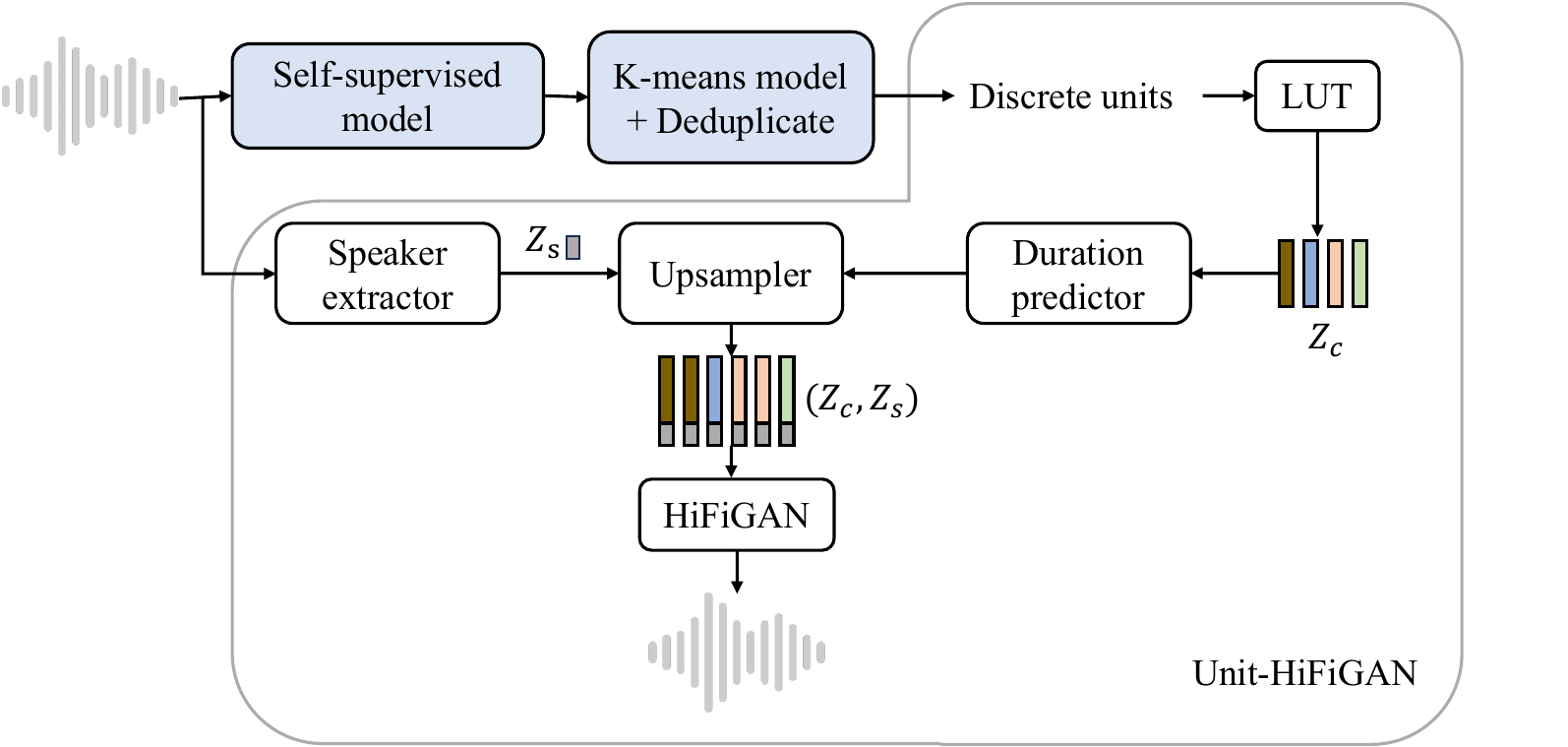}
  \caption{Architecture of speech re-synthesis system.}
  \vspace{-1em}
  \label{fig:unithifigan}
\end{figure}

Speech re-synthesis offers a more holistic evaluation of the representational capabilities of SSL models. The discrete units derived from SSL models can readily serve as inputs of speech re-synthesis and open the potential for advances of SLM. Therefore, we propose to investigate the performance on speech re-synthesis task. The speech re-synthesis system, as shown in Figure~\ref{fig:unithifigan}, is comprised of a SSL model based discrete unit extractor and a unit-HiFiGAN vocoder\cite{resynthesis}.

\begin{itemize}
    \item Discrete unit extractor. Discrete unit extractor aims to discretize representations extracted by self-supervised model into cluster indices using K-means model. The sequence of cluster indices are de-duplicated and serve as discrete units.
    \item Unit-HiFiGAN. A neural vocoder is employed to transform speech discrete units back into speech signal. As shown in Figure~\ref{fig:unithifigan}, as an input, the discrete units are converted into continuous representations $Z_c$ via look-up tables (LUT). Speaker embedding $Z_s$ is extracted from wavlm\cite{wavlm}, which is trained for speaker verification task, to support multi-speaker speech re-synthesis. The sequences $Z_c$ and $Z_s$ are then up-sampled based on the duration predicted by duration predictor. Speaker embedding $Z_s$ is concatenated to each frame of up-sampled embedding $Z_c$. Then, the concatenated sequence is transformed into waveform by the generator of HiFiGAN. 
\end{itemize}

\section{Experimental setup}
We evaluate the efficacy of our adaptation methods on adapting mHuBERT\cite{mhubert} to Mandarin, a challenging tonal language. HuBERT is known for its ability of discretizing speech signals into discrete units, and have been widely used in speech language models, such as SpeechLM\cite{speechlm}, SpeechGPT\cite{speechgpt} and SpeechGen\cite{speechgen}. The analysis of WER between transcripts of synthesized speech and ground truth ones serves a robust metric for measuring the performance of our adaptation methods and preservation strategies. 

In the two-iteration adaptation strategy, the target label of initial iteration is the clustering indices of 39-dimensional MFCC features. Subsequently, the representations extracted from 6th-layer of the model, obtained in first iteration, are clustered to form the target label of second iteration. In one-iteration adaptation strategy, the representations are extracted from 11-th layer of released model. The initial iteration in two-iteration adaptation strategy is trained for 250k steps. To have a fair comparison between two adaptation strategies, the second iteration is trained for 400k steps same as the one-iteration strategy.

 We use MiniBatchKMeans algorithm to realize K-means clustering. The mini-batch size is set to 10,000 frames. K-means++ with 20 random starts is used for better initialization. We set the rank $r$ in the adaptation methods to be 24 and integrate LoRA to the projection layer of query, key, value and output of self-attention layer in each Transformer block.

During adaptation, AISHELL-2\cite{aishell2} is employed, comprising 1000 hours of clean Mandarin reading-speech data. The English speech data used in preservation strategies is LibriSpeech\cite{librispeech}, from which a subset of 100 hours are randomly selected. The speech re-synthesis task is conducted using AISHELL-3\cite{aishell3}, an 85-hour high-fidelity multi-speaker Mandarin speech corpus designed for Text-to-Speech (TTS). Regarding the English corpus for speech re-synthesis, VCTK\cite{VCTK} is chosen, encompassing about 44-hour speech uttered by 110 English speakers. From the VCTK corpus, a random selection of 5\% of speakers constitutes the test set, while 95\% of the remaining speech from other speakers form the training set, with the rest allocated as the validation set.  The speech samples are uniformly down-sampled to \SI{16}{\kilo\hertz}.
\begin{table*}[t]
\centering
\caption{Effectiveness of adaptation methods on Mandarin speech re-synthesis task}
\vspace{-0.8em}
\label{tab: PEA effectiveness}
\begin{tabular}{l|l|c|c|c}
\hline
\multicolumn{1}{c|}{System}                                          & \multicolumn{1}{c|}{Size} & \multicolumn{1}{c|}{Trainable param.} & \multicolumn{1}{c|}{WER$\downarrow$} & MOS$\uparrow$       \\ \hline
Ground Truth                                                         & \multicolumn{1}{c|}{-}    & -                                         & 8.02\%  &4.58$\pm$0.07          \\ \hline
Mandarin HuBERT & 94.8 M & - & 26.79\%  & 3.27$\pm$0.09\\ \hline
\textbf{Before adaptation} &  & & \\
\hspace{2em}Un-adapted mHuBERT + Un-trained K-means                               & 94.8 M                      & -                                     & 50.57\% &2.26$\pm$0.11         \\

\hspace{2em}Un-adapted mHuBERT + Trained K-means                                 & 94.8 M                      & -                                      & 41.01\%   &2.55$\pm$0.12       \\ \hline
\textbf{After adaptation} & & & \\
\hspace{2em}Adapted mHuBERT (One-iteration) + Trained K-means  & 96.6 M                      & 2.026 M (2.098\% )                                  & 32.88\%   & 2.77$\pm$0.12       \\
\hspace{2em}Adapted mHuBERT (Two-iteration) + Trained K-means  & 96.6 M                      & 2.026 M (2.098\% )                                  & \textbf{21.94\%} & \textbf{3.80$\pm$0.08}\\ \hline
\end{tabular}
\vspace{-1.5em}
\end{table*}

\section{Results}

\subsection{Results on Mandarin adaptation}
 We conduct an analysis of WER and administer a Mean Opinion Score (MOS) test on the generated speech to evaluate the effectiveness. The transcripts of generated speech are generated by Whisper large-v3\cite{whisper}. For all MOS tests in this paper, 15 listeners are invited to assess 10 samples randomly selected from each system based on their naturalness. A reduction in WER and an increase in MOS after adaptation would suggest that the adapted model is capable of effectively encoding and tokenizing the new language, thereby serving as a robust indicator of successful adaptation. 
 
 As shown in "Before adaptation"  of Table \ref{tab: PEA effectiveness}, training K-means model without adaptation yields in an 18.90\% relative reduction in WER, suggesting that improved tokenization enhances performance. Applying  one-iteration adaptation strategy and two-iteration adaptation strategy to SSL model can achieve relative WER reductions of 34.98\% and 56.61\% respectively. This implies a potential trade-off between time-efficiency and adaptation performance, as two-iteration strategy gives superior target labels. Furthermore, Table \ref{tab: PEA effectiveness} reveals the parameter efficiency of our adaptation methods with less than 3\% of model parameters trainable. The results of MOS corroborate the WER findings, thereby validating our methods' efficacy. Additionally, a widely used open-sourced implementation of Madarin HuBERT trained on 1W+ hours of Wenetspeech\cite{wenetspeech} exhibits marginally inferior performance compared to our two-iteration adaptation strategy, which may be due to the differences in data distribution and  suggest our adaptation strategies can transfer knowledge from other languages. 
\subsection{Results on language preservation and adaptation}

To assess the extent of catastrophic forgetting, we perform a language maintenance test, depicted in "Before preservation" part of Table \ref{tab: bilingual adaptation}. In this test, the adapted mHuBERT model is utilized to extract English speech representations whereas the K-means model is trained on representations derived from un-adapted mHuBERT. English unit-HiFIGAN is utilized to transform speech units into waveform. It is observed that one-iteration adaptation strategy exhibits superior performance on existed language and after adaptation, there is a noticeable decline in performance on existed languages. 
\begin{table}[h]
\centering
\vspace{-0.5em}
\caption{Effectiveness of preservation strategies on English (en) and Mandarin (zh). In the system notion, the term preceding the hyphen denotes the language used during adaptation, whereas the term following it indicates the adaptation strategy.}
\vspace{-0.5em}
\label{tab: bilingual adaptation}
\resizebox{\linewidth}{!}
{
\setlength \tabcolsep{3pt}
\begin{tabular}{lcccc}
\hline
System                  & en WER$\downarrow$ & zh WER$\downarrow$ & en MOS$\uparrow$ & zh MOS$\uparrow$    \\ \hline

GT           & 1.44\%          & 8.02\%  &4.63$\pm$0.07 & 4.58$\pm$0.07        \\ \hline
English HuBERT           & 11.48\%          & - &4.03$\pm$0.08 &  -        \\ \hline
\multicolumn{5}{l}{\textbf{Before preservation}} \\ 
\hspace{0.2em} zh-2iter &38.17\% &\textbf{21.94}\%& 2.84$\pm$0.10   &  \textbf{3.80$\pm$0.08} \\
\hspace{0.2em} zh-1iter &15.50\% &32.88\%    & 3.97$\pm$0.08 & 2.77$\pm$0.12  \\ \hline
\multicolumn{5}{l}{\textbf{After preservation}}\\ 
\hspace{0.2em}zh-2iter   &13.84\% &23.71\%   & 4.04$\pm$0.08  & 3.61$\pm$0.08\\
\hspace{0.2em}zh-1iter   &9.70\% &34.36\%    & \textbf{4.24$\pm$0.07}  &2.80$\pm$0.10\\
\hspace{0.2em}enzh-2iter &\textbf{7.94}\% &27.18\%   &4.21$\pm$0.08    &3.35$\pm$0.10
     \\ \hline
\end{tabular}
}
\vspace{-0.5em}
\end{table}

The effectiveness of preservation strategies are shown in Table \ref{tab: bilingual adaptation}. Speech units of Mandarin and English are converted using bi-lingual unit-HiFiGAN. It is observed that employing solely Mandarin speech for adaptation, one-iteration adaptation strategy yields better results in English but worse results in Mandarin. This phenomenon underscores the delicate balance between adapting a new language and preserving existed languages. Subsequent to re-clustering with combined data, there is a marked improvement in the performance of existed language, with WER reducing from 38.17\% to 13.84\% in two-iteration adaptation strategy. Besides, our method behaves even better with the original model (before any adaptation) in existed language with the performance in new language remains largely unaffected. These results confirmed the performance of our preservation strategies. Discrepancies between WER and MOS are noted, particularly when WER values are close, as discerning differences between two similarly performing systems may be challenging for listeners. Nevertheless, after adaptation, there is a significant increase in MOS for English, rising from 2.84 to 4.04 in two-iteration adaptation strategy. The strategy of combining data during adaptation has been shown to improve performance in existed languages while reducing more performance on new language. The impressive performance of adapted mHuBERT, utilizing speech exclusively from new language and employing re-clustering preservation strategy, facilities the processing of both Mandarin and English effectively.

\subsection{Ablation study}
We also enlarge the clustering number of K-means model to 3000 to examine the impact of clustering number. For Adaptation, we only focus on the performance of new language, whereas for preservation, we consider the balance between new languages and existed language. It is observed from Table \ref{tab: ablation study} that increasing the number of clusters yields the most substantial adaptation performance, with a relative WER reduction of up to 61.72\% compared to un-adapted model. During preservation, the performance enhancement in both existed and new languages can also be observed, with WER reduced up to around 2\% in each language.
\begin{table}[h]
\setlength \tabcolsep{12pt}
\centering
\vspace{-0.3em}
\caption{Ablation study on cluster numbers}
\vspace{-1em}
\label{tab: ablation study}
\begin{tabular}{lcc}
\hline
System                     & en WER$\downarrow$     & zh WER$\downarrow$  \\ \hline
\hline
\textbf{Adaptation} &            &         \\
\hspace{2em}zh-2iter-1000              & -          & 21.94\% \\
\hspace{2em}zh-2iter-3000              & \textbf{-} & 19.36\% \\ \hline
\hline
\textbf{Preservation} &            &         \\
\hspace{2em}zh-2iter-1000              & 13.84\%    & 23.71\% \\
\hspace{2em}zh-2iter-3000              & 12.53\%    & 21.5\%  \\ \hline
\end{tabular}
\vspace{-1.4em}
\end{table}
\section{Discussion}

In this paper, we introduce parameter efficient adaptation methods which integrate LoRA into existed SSL models and preservation strategies including re-clustering and data combination to adapt an existed SSL model to a new language without harming its original abilities. We select speech re-synthesis task to evaluate the efficacy of our proposed approaches. Experiments corroborate that our methods can seamlessly improve the multilingual mastery in SSL models while preserving capabilities for existed languages. Currently, we only focus on a most-widely adopted and representative SSL model, mHuBERT. In the future, we plan to assess the effectiveness of our methods across a broader range of SSL models and diverse downstream tasks.

\section{Acknowledgements}

The first author Jing Xu would like to thank the guidance and support from Xiao Chen, Daxin Tan and Yuting Yeung. This work is partially supported by the Human-Computer Communications Laboratory, Department of Systems Engineering \& Engineering Management, The Chinese University of Hong Kong.

\bibliographystyle{IEEEtran}
\bibliography{mybib}

\end{document}